%% file: main.tex
\documentclass[10pt,twocolumn,letterpaper]{article}

\usepackage{cvpr}              
\usepackage[ruled,vlined,linesnumbered]{algorithm2e}
\usepackage{amsmath,amssymb}
\usepackage{graphicx}
\usepackage{svg}
\usepackage{multirow}
\usepackage[table,xcdraw]{xcolor}

\input{preamble}

\definecolor{cvprblue}{rgb}{0.21,0.49,0.74}
\usepackage[pagebackref,breaklinks,colorlinks,allcolors=cvprblue]{hyperref}

\def\paperID{***} 
\def\confName{CVPR}
\def\confYear{2026}

\title{Multi-modal Test-time Adaptation via Adaptive Probabilistic Gaussian Calibration}

\author{
Jinglin Xu$^{*12}$ 
\quad
Yi Li$^{*12}$
\quad
Chuxiong Sun$^{1}$
\quad
Xiao Xu$^{3}$
\quad
Jiangmeng Li$^{\dagger 1}$
\quad
Fanjiang Xu$^{1}$\\[1mm]
$^{1}$Institute of Software Chinese Academy of Sciences\\
$^{2}$University of Chinese Academy of Sciences\\
$^{3}$National Defense University\\[1mm]
{\tt\small \{xujinglin2024,liyi2022,chuxiong2016,jiangmeng2019,fanjiang\}@iscas.ac.cn,
xuxiao0825@gmail.com
}
}

\begin{document}
\maketitle
\begingroup
\renewcommand\thefootnote{*}
\footnotetext{Equal contribution.}
\renewcommand\thefootnote{$\dagger$}
\footnotetext{Corresponding author.}
\endgroup
\input{sec/0_abstract}    
\input{sec/1_intro}

\input{sec/2_related}

\input{sec/3_method}

\input{sec/4_exp}

\input{sec/5_conclusion}

{
    \small
    \bibliographystyle{ieeenat_fullname}
    \bibliography{main}
}

\end{document}

%% file: preamble.tex
\usepackage{adjustbox}



%% file: sec/0_abstract.tex
\begin{abstract}
Multi-modal test-time adaptation (TTA) enhances the resilience of benchmark multi-modal models against distribution shifts by leveraging the unlabeled target data during inference. Despite the documented success, the advancement of multi-modal TTA methodologies has been impeded by a persistent limitation, i.e., the lack of explicit modeling of category-conditional distributions, which is crucial for yielding accurate predictions and reliable decision boundaries. Canonical Gaussian discriminant analysis (GDA) provides a vanilla modeling of category-conditional distributions and achieves moderate advancement in uni-modal contexts. However, in multi-modal TTA scenario, the inherent modality distribution asymmetry undermines the effectiveness of modeling the category‑conditional distribution via the canonical GDA. To this end, we introduce a tailored probabilistic Gaussian model for multi-modal TTA to explicitly model the category-conditional distributions, and further propose an adaptive contrastive asymmetry rectification technique to counteract the adverse effects arising from modality asymmetry, thereby deriving calibrated predictions and reliable decision boundaries. Extensive experiments across diverse benchmarks demonstrate that our method achieves state-of-the-art performance under a wide range of distribution shifts. The code is available at https://github.com/XuJinglinn/AdaPGC.
\end{abstract}

%% file: sec/1_intro.tex
\section{Introduction}
\label{sec:intro}

Multi-modal learning, which integrates information from heterogeneous data sources, has demonstrated considerable promise across diverse application domains, including human action recognition \cite{das2020mmhar,liu2020benchmark}, sentiment analysis \cite{yu2022unified,yu2021learning}, and other complex perception tasks \cite{xin2024mmap,zhang2024tamm,kamath2021mdetr}. Conventional multi-modal models are often vulnerable to distribution shifts, wherein discrepancies between the source and target domains can precipitate pronounced performance degradation.
In response to this limitation, multi-modal test-time adaptation (TTA) \cite{shin2022mm,DBLP:conf/iclr/Yang0Z0024} has emerged as a viable paradigm, which dynamically adapts the multi-modal models using only unlabeled target data, thereby facilitating real-time robustness enhancement under distribution shifts.

\begin{figure}[t]
  \centering
  \includegraphics[width=0.99\linewidth]{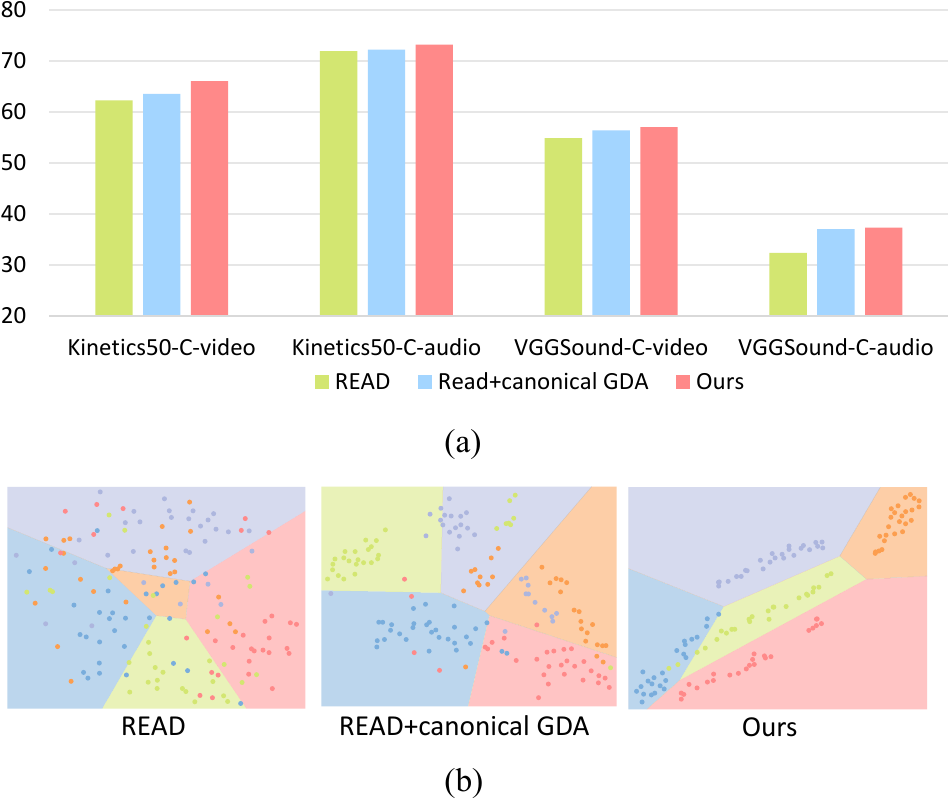}
  \caption{READ \cite{DBLP:conf/iclr/Yang0Z0024} is a representative multi-modal TTA benchmark. Both READ+canonical GDA and our method explicitly model the category-conditional distributions. 
  (a) We compare the prediction accuracy of three methods on Kinetics50-C and VGGSound-C datasets, with corruption applied to either the video or audio data.
  (b) We plot the decision boundaries of three methods on Kinetics50-C-video, where points of different colors represent different categories.}
  \label{fig:motivation experiment}
\end{figure}

Nevertheless, relying on black-box neural networks to derive task-dependent predictions, existing multi-modal TTA methods generally overlook the explicit modeling of category-conditional distributions.
As the empirical results in Figure \ref{fig:motivation experiment} show, the multi-modal TTA benchmark lacks the explicit modeling of category-conditional distributions, thus achieving inferior prediction accuracy and irregular boundaries. 
In contrast, the methods with explicit modeling of category-conditional distributions yield better predictions and smoother boundaries.
As a result, the explicit modeling of category-conditional distributions, neglected by existing multi-modal TTA benchmarks, is crucial for obtaining reliable decision boundaries.

Recent uni-modal methods \cite{cui2025bayestta, han2024dota} provide a vanilla modeling of category-conditional distributions by the canonical Gaussian discriminant analysis (GDA) \cite{liang2022gmmseg}, and achieve moderate advancement.
Yet, in the multi-modal TTA scenarios, the inherent \textit{modality distribution asymmetry} undermines the effectiveness of utilizing the canonical GDA for modeling the category-conditional distributions. \textit{Modality distribution asymmetry} refers to the phenomenon that the collected multi-modal data do not necessarily exhibit distributional shifts in each modality, only one particular modality may experience such a shift instead. This asymmetry arises because many real-world corrupting factors affect only specific modalities.\footnote{For example, an autonomous driving system typically needs to integrate data from multiple modalities (such as camera, LiDAR, and radar) to make decisions, the damp ground after the rain primarily impacts the LiDAR signals, while the night scenes mainly affect the camera signals.}
The canonical GDA approach models category-conditional distributions based on category-specific means and covariances \cite{DBLP:conf/icml/CaoXDZH24}. Treating corrupted modality and normal modality equally can lead to biased estimation of the means, and distribution shifts affect the dispersion of features within the same category \cite{zhang2023provable,DBLP:conf/iclr/Li0Z0XY25}. Therefore, \textit{modality distribution asymmetry} limits the effectiveness of canonical GDA in multi-modal TTA scenarios, and the results in Figure \ref{fig:motivation experiment} demonstrate our conclusion.

To this end, we introduce an \textit{\textbf{Ada}ptive \textbf{P}robabilistic \textbf{G}aussian \textbf{C}alibration} (AdaPGC) method for multi-modal TTA. AdaPGC consists of two modules: probabilistic Gaussian prediction calibration and adaptive contrastive asymmetry rectification. Specifically, in the probabilistic Gaussian prediction calibration module, we leverage the probabilistic Gaussian model to explicitly model category-conditional distributions, and progressively update category means and
covariance matrix without relying on supervision or source data. In the adaptive contrastive asymmetry rectification module, we adaptively detect which modality has been corrupted and employ a contrastive rectification approach to counteract the adverse effects arising from modality distribution asymmetry. Overall, our contribution can be summarized as follows:
\begin{itemize}
    \item We empirically demonstrate that the explicit modeling of category-conditional distributions is vital for calibrated predictions and reliable decision boundaries, and further reveal that \textit{modality distribution asymmetry} challenges the effective explicit modeling of category-conditional distributions in the scenario of multi-modal TTA.

    \item To address the aforementioned limitations, we propose an adaptive probabilistic Gaussian calibration method for multi-modal TTA (dubbed as AdaPGC), which consists of probabilistic Gaussian prediction calibration and adaptive contrastive asymmetry rectification.

    \item We conduct comprehensive experiments on two widespread multi-modal TTA benchmarks, and the results show that AdaPGC achieves the state-of-the-art performance. The thorough experimental setup guarantees the robustness and generalizability of AdaPGC.
    
\end{itemize}

%% file: sec/2_related.tex
\section{Related work}

\subsection{Transition from TTA to Multi-modal TTA }
Test-time adaptation aims to adapt a pre-trained model to the target domain without access to the source data. 
The challenge lies in leveraging only the unlabeled target data to overcome the distribution shifts between source and target domains.  
Early test-time adaptation (TTA) methods primarily focused on single-modal TTA. TENT \cite{wang2021tent} optimizes the normalization layers by minimizing the prediction entropy of the model to reduce the distribution shifts between source and target domains; 
EATA \cite{pmlr-v162-niu22a} and SAR \cite{niu2023towards} employ sample selection strategies to facilitate the adaptation of the model by using more informative target samples; T3A \cite{iwasawa2021test} improves performance by utilizing off-the-shelf target data and leveraging external knowledge about the target domain.

Compared with uni-modal TTA, multi-modal TTA is more challenging because different modalities may suffer from different corruption levels, which can significantly degrade performance. 
MM-TTA \cite{shin2022mm} mitigates this issue through complementary pseudo-labeling across modalities, but its robustness still depends on the quality of the generated pseudo-labels.
In contrast, READ \cite{DBLP:conf/iclr/Yang0Z0024} introduces a more robust approach by enhancing the impact of confident predictions and reducing the influence of noisy predictions. 
Meanwhile, ABPEM \cite{zhao2025attention} improves multi-modal TTA by mitigating modality misalignment under distribution shift through attention bootstrapping and principal entropy minimization.
More recently, SuMi \cite{guo2025smoothing} improves multi-modal TTA robustness under mixed shifts, multi-modal corruptions, and missing modalities through smoothing and cross-modal information sharing.
Recent work TSA \cite{chen2025testtime} further explores uni-modal distribution shift problem via a selective adaptation framework to handle shifts in the affected modality.
However, existing multi-modal TTA methods overlook the explicit modeling of category-conditional distributions, which limits their ability to obtain accurate predictions and reliable decision boundaries.

\subsection{Test-Time Adaptation Methods Based on GDA}

Gaussian discriminant analysis (GDA) is a method capable of modeling category-conditional distributions. 
Unlike prototype-based approaches that only compute distances to fixed category prototypes, GDA captures the internal distributional structure of each class, enabling more precise adaptation to distribution shifts. 
GDA has demonstrated promising empirical performance in uni-modal TTA.

DOTA \cite{han2024dota} incorporates GDA to continuously estimate the underlying distribution of the stream target data. 
BayesTTA \cite{cui2025bayestta} enhances GDA by dynamically selecting covariance structures and enforcing temporal consistency, thereby improving the representations under evolving distributions. 
ADAPT \cite{zhangbackpropagation} introduces a historical knowledge bank to correct potential likelihood biases, further enhancing GDA.
While these methods achieve empirical gains by leveraging the canonical GDA to explicitly model the category-conditional distributions in the uni-modal context, the inherent \textit{modality distribution asymmetry} hinders the effectiveness of the canonical GDA in the multi-modal scenario.

%% file: sec/3_method.tex
\section{Preliminary}
In this section, we present the basic preliminaries about the canonical GDA, which is a classical generative model that assigns category labels based on the likelihood of category-conditional distributions. Specifically, considering a batch of samples $\{(\boldsymbol{x}_i,y_i)\}_{i=1}^{\mathcal{B}}$, where $\mathcal{B}$ is the batch size and $y_i \in \{1,2,\cdots,C\}$ ($C$ is the number of categories), the canonical GDA assumes that features conditioned on category $c$ ($c=1,\dots,C$) follow a Gaussian distribution with a shared covariance matrix:
\begin{equation}
\begin{aligned}
    p_{i,c} & = p(\boldsymbol{x}_i|y_i=c) = \mathcal{N}(\boldsymbol{x}_i; \boldsymbol{\mu}_c, \boldsymbol{\Sigma}) \\
    & = \frac{1}{\sqrt{(2\pi)^d |\boldsymbol{\Sigma}|}}
\exp\left( -\frac{1}{2} (\boldsymbol{x}_i - \boldsymbol{\mu}_c)^\top \Sigma^{-1} (\boldsymbol{x}_i - \boldsymbol{\mu}_c) \right),
\end{aligned}
\end{equation}
where $\boldsymbol{\mu}_c$ and $\boldsymbol{\Sigma}$ denote the category mean and shared covariance matrix, respectively.

From the Bayes’ theorem, the posterior probability of the category $c$ given sample $\boldsymbol{x}_i$ is
$
p(c|\boldsymbol{x}_i)= \frac{p(c) \cdot p(\boldsymbol{x}_i|c)}{p(\boldsymbol{x}_i)} \propto \pi_c \; \mathcal{N}(\boldsymbol{x}_i;\boldsymbol{\mu}_c, \boldsymbol{\Sigma}),
$
where $\pi_c$ is the category prior. The canonical GDA assumes a uniform category prior (i.e., $\pi_c = \frac{1}{C}$), then the Bayes optimal prediction label for $\boldsymbol{x}_i$ is known as the $\arg\max$ of $\hat{y}_{i,c}$ over $c=1,\dots,C$:
\begin{equation}
\hat{y}_{i,c} = \boldsymbol{w}_c^\top \boldsymbol{x}_i + b_c,
\end{equation} where $\boldsymbol{w}_c = \boldsymbol{\Sigma}^{-1} \boldsymbol{\mu}_c,\quad 
\boldsymbol{b}_c = -\frac{1}{2} \boldsymbol{\mu}_c^\top \Sigma^{-1} \boldsymbol{\mu}_c$.
The estimation of $\boldsymbol{\mu}_c$ and $\boldsymbol{\Sigma}$ can be optimization-free, since their maximum likelihood estimators under the i.i.d.\ assumption are just the category-wise empirical averages and the pooled sample covariance matrix of $\boldsymbol{x}_i$.

\section{Method}

\begin{figure*}[t]
  \centering
  \includegraphics[width=\linewidth]{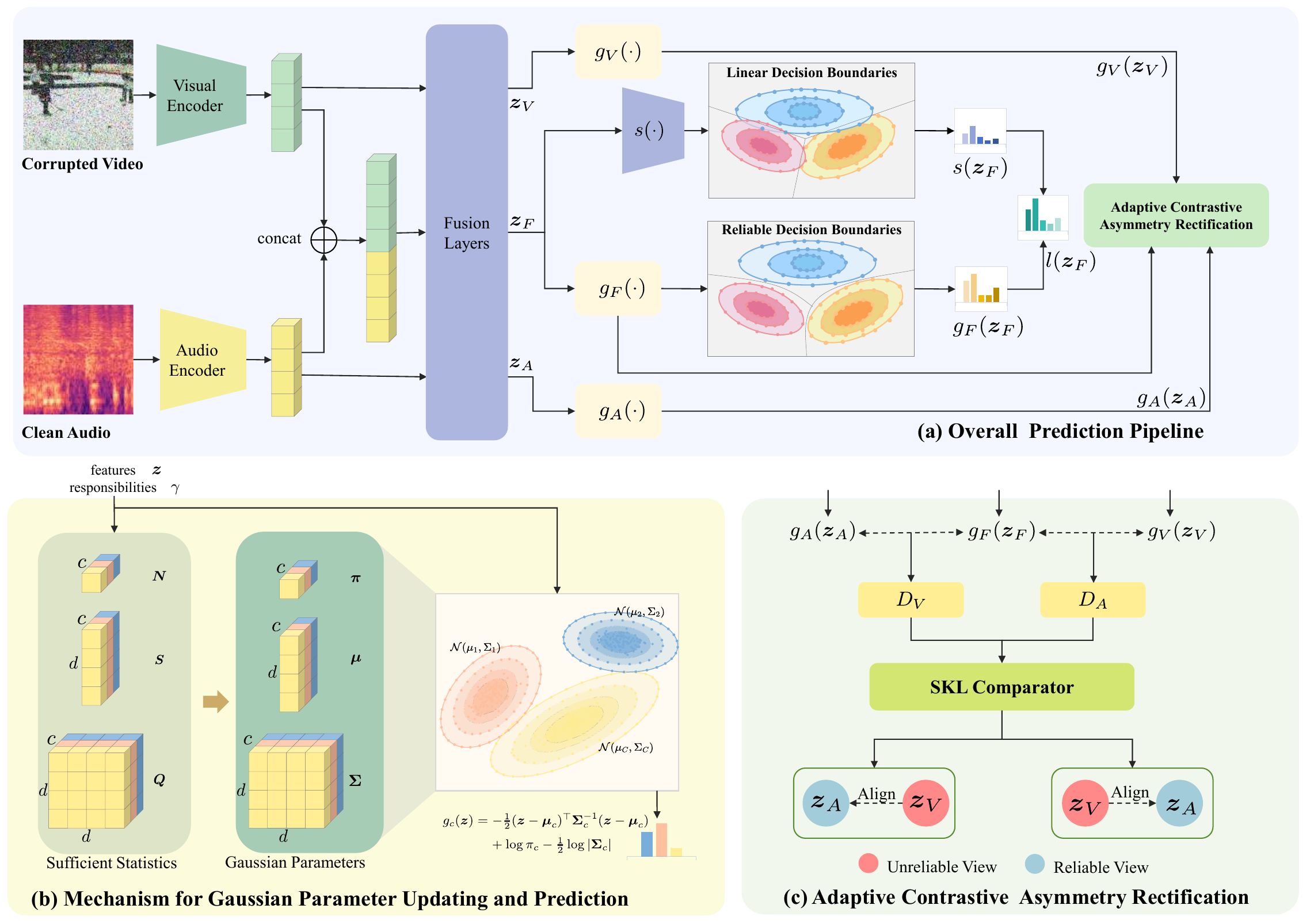}
\caption{Overview of the proposed multi-modal test-time adaptation framework.
(a) Overall Prediction Pipeline.
Video and audio inputs are encoded into modality-specific features and concatenated to form a full-modality representation. Each branch is mapped by a scoring function $g(\cdot)$ to category-posterior estimates, which support prediction and modality-reliability assessment; the detailed formulation of $g(\cdot)$ is provided in panel (b).
(b) Mechanism for Gaussian Parameter Updating and Prediction.
Streaming target data progressively update the sufficient statistics $(N,S,Q)$, from which category priors, means, and covariance matrices are incrementally estimated to maintain adaptive class-conditional density models for prediction.
(c) Adaptive contrastive asymmetry rectification.
Posterior consistency across modalities is evaluated using symmetric KL divergence to identify the reliable view. A one-sided alignment then pulls the unreliable feature toward the reliable one while keeping the reliable branch fixed.}
  \label{fig:arch}
\end{figure*}

Figure~\ref{fig:arch} presents the overall framework of AdaPGC. In the following, we describe each component in detail.

\subsection{Problem formulation}
Without loss of generality, we take two modalities (i.e., $m_1$ and $m_2$) as an example for clarity of presentation. Let $\mathcal{M}_{\theta}=(\phi_{1},\phi_2,\mathcal{F})$ be the multi-modal model, $\theta$ is the parameter to be adapted at test-time, $\phi_{1},\phi_2$ are the encoders of $m_1,m_2$ modalities, and $\mathcal{F}$ is the multi-modal fusion layers. The target data arrives in the form of a stream: $\mathcal{D}_{target}=\{\boldsymbol{x}_i^t\}_{i=1}^{\mathcal{B}}$, where $\boldsymbol{x}_i^t=\{\boldsymbol{x}_{m_1,i}^t,\boldsymbol{x}_{m_2,i}^t\}$, $t \in [0,T]$ is the index of batch and $\mathcal{B}$ is the batch size. Existing multi-modal TTA methods generally update the parameter $\theta$ based on the minimization of target data's entropy:
\begin{equation}
    \mathrm{Ent}_\theta(\boldsymbol{x})=-\boldsymbol{p}_\theta(\boldsymbol{x})\log\boldsymbol{p}_\theta(\boldsymbol{x})=-\sum_{i=1}^Cp_\theta(\boldsymbol{x})_i\log p_\theta(\boldsymbol{x})_i,
\end{equation}
where $\boldsymbol{p}_{\theta}(\boldsymbol{x}) = \left(p_{\theta}(\boldsymbol{x})_1, p_{\theta}(\boldsymbol{x})_2, \cdots , p_{\theta}(\boldsymbol{x})_C \right)= \text{Softmax}(\mathcal{M}_{\theta}(\boldsymbol{x})) $ is the probabilistic distribution
outputted by the model $\mathcal{M}_{\theta}$ and $C$ is the number of categories.

\subsection{Probabilistic Gaussian prediction calibration}
As mentioned in \textbf{Section} \ref{sec:intro}, due to the lack of explicit modeling of the category‑conditional distributions, the predictions of existing multi‑modal TTA methods often exhibit sub‑optimal performance. 
In this subsection, we elaborate on the formulation of how a probabilistic Gaussian model is employed to characterize the category‑conditional distributions in multi‑modal TTA and to calibrate the predictions.

In practical applications, inter‑category differences manifest in more than mere shifts of the mean \cite{cui2025bayestta}. 
Consequently, the assumption in the canonical GDA that the samples from different categories share an identical covariance matrix is impractical, and we replace the identical covariance matrix by category-specific covariance matrix. 
Meanwhile, multi-modal TTA operates without supervision on the target domain and lacks access to source data, which renders the estimation of mean, covariance matrix and prior of each category challenging, thus we choose to progressively update the mean, covariance matrix and prior of each category by the streaming target data.

We model the category-conditional distributions by the penultimate-layer features (i.e., the representations right before the final linear classifier).
For each test sample $\boldsymbol{x}_{i}^t$, we obtain three $d$-dimensional features: $
\boldsymbol{z}_{m_1,i}^t,
\boldsymbol{z}_{m_2,i}^t,
\boldsymbol{z}_{F,i}^t,$
which correspond to two uni-modal features $m_1,m_2$ and one multi-modal feature $F$.

For each perspective $m \in \{m_1,m_2,F\}$ and category $c$, we assume a Gaussian category-conditional distribution
\begin{equation}
  p(\boldsymbol{z}_{m,i}^t \mid y_i = c)
  = \mathcal{N}\!\big(\boldsymbol{z}_{m,i}^t;\,\boldsymbol{\mu}_{m,c}^t,\boldsymbol{\Sigma}_{m,c}^t\big),
\end{equation}
where $\boldsymbol{\mu}_{m,c}^t$ and $\boldsymbol{\Sigma}_{m,c}^t$ denote the category-specific mean and covariance of perspective $m$ estimated from the $t$-th batch.

From the law of total probability, the marginal likelihood of a feature is given by
$p(\boldsymbol{z}_{m,i}^t)=\sum_{c=1}^C \pi_{m,c}^t\,p(\boldsymbol{z}_{m,i}^t\mid c)$
where $\pi_{m,c}^t$ denotes the category prior. 
The posterior distribution follows directly from Bayes’ theorem:
\begin{equation} p(c\mid \boldsymbol{z}_{m,i}^t) =\frac{\pi_{m,c}^t\,p(\boldsymbol{z}_{m,i}^t\mid c)}{\sum_{c'}\pi_{m,c'}^t\,p(\boldsymbol{z}_{m,i}^t\mid c')}. 
\end{equation}

Up to a category-independent constant, the log-posterior is
\begin{equation}
\label{eq:gda-score}
\begin{aligned}
g_{m,c}(\boldsymbol{z}_{m,i}^t) &= \log p(c\mid \boldsymbol{z}_{m,i}^t)\\
&= -\tfrac12 (\boldsymbol{z}_{m,i}^t-\boldsymbol{\mu}_{m,c}^t)^\top \boldsymbol{\Sigma}_c^{-1}(\boldsymbol{z}_{m,i}^t-\boldsymbol{\mu}_{m,c}^t)
    \\
&\quad + \log \pi_{m,c}^t - \tfrac12 \log|\boldsymbol{\Sigma}_{m,c}^t|+\text{const}.
\end{aligned}
\end{equation}
The constant term is irrelevant to category comparison and may be omitted when convenient.

Hence, every perspective $m$ has its own log-posterior function $g_{m,c}(\cdot)$ derived from its posterior. In the sequel, we use $g_m(\cdot)$ to denote the vector of category scores produced by view $m$.

Inspired by \cite{iwasawa2021test}, which distills usable templates from a source model, we initialize the parameters $\boldsymbol{\mu}_{m,c}^0,\boldsymbol{\Sigma}_{m,c}^0$ and $\pi_{m,c}^0$ directly from the source model’s linear classification head.

Let the source model’s final linear score be
\begin{equation}
s_c(\boldsymbol{z})=\boldsymbol{w}_c^\top \boldsymbol{z} + b_c,
\end{equation}
which corresponds to the output of its last linear classification layer.
Assume the initial value of the $\boldsymbol{\mu}_{m,c}^0$ is linear weights $\boldsymbol{w}_c$  and the initial state of the covariance is the identity matrix $I$. 
Substituting it into the log-posterior \cref{eq:gda-score}  gives
\begin{align}
g_{m,c}(\boldsymbol{z}_{m,i}^0) = \boldsymbol{w}_c^\top \boldsymbol{z}_{m,i}^0 - \tfrac12\|\boldsymbol{w}_c\|^2 + \log \pi_{m,c}^0- \tfrac12 \log |\boldsymbol{\Sigma}_{m,c}^0|.
\end{align}
By choosing $\log \pi_{m,c}^0 = b_c + \tfrac12 \|\boldsymbol{w}_c\|^2 + \tfrac12\log |\boldsymbol{\Sigma}_{m,c}^0|,$ the resulting log-posterior and the source model’s linear score satisfy $g_c(\boldsymbol{z})=s_c(\boldsymbol{z})+\text{const}$, where the additive constant is identical for all classes. In particular, when using the default covariance $\boldsymbol{\Sigma}_c=I$,
\begin{align}
\boldsymbol{\mu}_{m,c}^0 & =\boldsymbol{w}_c,\\ 
\boldsymbol{\Sigma}_{m,c}^0 & =I,\\ 
\log\pi_{m,c}^0 & =b_c+\tfrac12\|\boldsymbol{w}_c\|^2\ + \tfrac12\log |\boldsymbol{\Sigma}_{m,c}^0|.
\end{align}
This absorbs the linear bias $b_c$ into the category prior $\pi_{m,c}^0$, ensuring that the initial log-posterior predictions are fully aligned with those of the source classifier.

In multi-modal TTA, target samples arrive sequentially and often differ significantly from the source-domain distribution. Consequently, the initial parameters $\boldsymbol{\mu}_{m,c}^0,\boldsymbol{\Sigma}_{m,c}^0$ and $\pi_{m,c}^0$ obtained from the pretrained source model may no longer be reliable.
To mitigate this mismatch, we progressively update these parameters with each incoming batch of target data, allowing the model to adapt to the evolving target distribution.

For each component $c$, we maintain total soft count $N_c^t$,
first-order sufficient statistic $\mathbf{S}_c^t \in \mathbb{R}^d$, and second-order sufficient
statistics $\mathbf{Q}_c^t \in \mathbb{R}^{d \times d}$.
\begin{equation*}
N_{m,c}, \qquad
\mathbf{S}_{m,c} = \sum_{i \in \mathcal{I}_c} \mathbf{z}_{m,i}, \qquad
\mathbf{Q}_{m,c} = \sum_{i \in \mathcal{I}_c} \mathbf{z}_{m,i}\mathbf{z}_{m,i}^{\top}
\end{equation*}
yields $\boldsymbol{\mu}_{m,c} = \mathbf{S}_{m,c}/N_{m,c}$ and
$\boldsymbol{\Sigma}_{m,c} = \mathbf{Q}_{m,c}/N_{m,c} - \boldsymbol{\mu}_{m,c}\boldsymbol{\mu}_{m,c}^{\top}$,
where $\mathcal{I}_c$ is the set of samples assigned to category $c$.
These quantities admit streaming updates without storing past samples, which is crucial for TTA.

Given a batch with responsibilities $\{\gamma_{ic}^t\}$, where each$\gamma_{ic}^t$ represents the soft category assignment produced by the source model for sample $i$ and category $c$, we update

$
\begin{aligned}
N_{m,c}^{t} &= N_{m,c}^{t-1} + \Delta N_{m,c}^{t}, &
\Delta N_{m,c}^{t} &= \sum_i \gamma_{ic}^{t}, \\
\mathbf{S}_{m,c}^{t} &= \mathbf{S}_{m,c}^{t-1} + \Delta \mathbf{S}_{m,c}^{t}, &
\Delta \mathbf{S}_{m,c}^{t} &= \sum_i \gamma_{ic}^{t}\,\boldsymbol{z}_{m,i}^t, \\
\mathbf{Q}_{m,c}^{t} &= \mathbf{Q}_{m,c}^{t-1} + \Delta \mathbf{Q}_{m,c}^{t}, &
\Delta \mathbf{Q}_{m,c}^{t} &= \sum_i \gamma_{ic}^{t}\,\boldsymbol{z}_{m,i}^t \boldsymbol{z}_{m,i}^{t\top}.
\end{aligned}
$
The moment MLEs (i.e., the maximum likelihood estimates of the class prior, mean, and covariance) are

\begin{equation}
\begin{aligned}
\hat{\pi}_{m,c}^t & = \frac{N_{m,c}^t}{\sum_c' N_{m,c'}^t},\\
\hat{\boldsymbol{\mu}}_{m,c}^t & =\, \frac{\mathbf{S}_{m,c}^t}{N_{m,c}^t},\\
\hat{\boldsymbol{\Sigma}}_{m,c}^t & =\, \frac{\mathbf{Q}_{m,c}^t}{N_{m,c}^t}\;-\;\hat{\boldsymbol{\mu}}_{m,c}^t\hat{\boldsymbol{\mu}}_{m,c}^{t\top}.
\end{aligned}
\end{equation}

To obtain smooth and numerically stable evolution, we apply EMA \cite{adam2014method} updates with rate $\alpha=0.9$:
\begin{equation}
\begin{aligned}
\pi_{m,c}^{t} &= \alpha\,\pi_{m,c}^{t-1} + (1-\alpha)\,\hat{\pi}_{m,c}^{t},\\
\boldsymbol{\mu}_{m,c}^{t} &= \alpha\,\boldsymbol{\mu}_{m,c}^{t-1} + (1-\alpha)\,\hat{\boldsymbol{\mu}}_{m,c}^{t},\\
\boldsymbol{\Sigma}_{m,c}^{t} &= \alpha\,\boldsymbol{\Sigma}_{m,c}^{t-1} + (1-\alpha)\,\hat{\boldsymbol{\Sigma}}_{m,c}^{t}.
\end{aligned}
\end{equation}

With the category-conditional parameters updated, the real-time prediction can then be obtained from \cref{eq:gda-score}. To integrate the source model’s original logits with the evidence provided by log-posterior, 
we form the fused logits
\begin{equation}
l(\boldsymbol{z}_{F,i})
= s(\boldsymbol{z}_{F,i}) + \lambda\,g_{F}(\boldsymbol{z}_{F,i}),
\label{eq:final-pre}
\end{equation}
where $s(\boldsymbol{z}_{F,i}), g_{F}(\boldsymbol{z}_{F,i}) \in \mathbb{R}^{C}$ denote the source logits and GDA scores on the fused view, respectively, and $\lambda$ is a fusion weight that controls the influence of the GDA evidence. The final prediction is $
\hat{y}_i = \arg\max_{c} l_c(\boldsymbol{z}_{F,i}).
$

However, direct summation without the correct constraints in place may cause the two scores to interfere with each other, resulting in confidence oscillations. 
To address this issue, we employ soft-prediction alignment to minimize distributional conflict before fusion.

Let
\begin{align}
        \boldsymbol{p}^{\text{lp}}(\boldsymbol{z}_{F,i})
 & = \text{Softmax}\big(g_{F}(\boldsymbol{z}_{F,i})\big),\\
\boldsymbol{p}^{\text{src}}(\boldsymbol{z}_{F,i})
 & = \text{Softmax}\big(s(\boldsymbol{z}_{F,i})\big)
\end{align}
denote the predictions from log-posterior and the source model, respectively. 
We define the alignment loss as
\begin{equation}
\mathcal{L}_{\text{g}}
= -\,\mathbb{E}\!\left[
\sum_{c=1}^{C}
p^{\text{lp}}_{c}(\boldsymbol{z}_{F,i})\,
\log p^{\text{src}}_{c}(\boldsymbol{z}_{F,i})
\right].
\label{eq:l-g}
\end{equation}
Here, $\boldsymbol{p}^{\text{lp}}(\boldsymbol{z}_{F,i})$ is treated as a fixed reference (with gradients detached), which gently guides the source branch decision surface toward the log-posterior and stabilizes the fused predictions.

\subsection{Adaptive contrastive asymmetry rectification}

As verified in \textbf{Section} \ref{sec:intro}, in the scenario of multi-modal TTA, 
the inherent \textit{modality distribution asymmetry} distorts the estimation of category mean and covariance matrix, thereby undermining the effectiveness of the probabilistic Gaussian prediction calibration.
To address this limitation, we adaptively detect which modality exhibits a distribution shift, and subsequently leverage contrastive rectification to counteract the adverse effects arising from \textit{modality distribution asymmetry}.

From the log-posterior, we obtain an alternative view of the prediction through $p(c\mid\boldsymbol{z}_{m,i}^t)=\text{Softmax}(g_{m,c}(\boldsymbol{z}_{m,i}^t))$.
As demonstrated in \cite{guo2025smoothing}, the multi-modal prediction $p(c\mid \boldsymbol{z}_{F,i}^t)$ is more robust and  stable than the uni-modal prediction $p(c\mid \boldsymbol{z}_{m_1,i}^t)$ or $p(c\mid \boldsymbol{z}_{m_2,i}^t)$. Therefore, we detect modalities exhibiting distribution shift by comparing the distribution discrepancies between different uni-modal predictions and multi-modal predictions. Specifically, we compute the distribution discrepancy by
\begin{equation}
\begin{aligned}
    & D_{m_1,i}^t=\mathrm{SKL}\big(p(c\mid\boldsymbol{z}_{m_1,i}^t),p(c\mid \boldsymbol{z}_{F,i}^t)\big), \\
    & D_{m_2,i}^t=\mathrm{SKL}\big(p(c\mid\boldsymbol{z}_{m_2,i}^t),p(c\mid \boldsymbol{z}_{F,i}^t)\big),
\end{aligned}
\end{equation}
where $\mathrm{SKL}(\cdot,\cdot)$ is the symmetric KL divergence calculated by
\begin{equation}
   \mathrm{SKL}(p,q)=\frac{1}{2} \Big( \mathrm{KL}(p\Vert q)+\mathrm{KL}(q\Vert p)\Big). 
\end{equation}

A modality that exhibits a larger distribution discrepancy from the multi-modal prediction is regarded as a modality with distribution shift. Based on this criterion, we partition the batch indices into two sets:
\begin{equation}
\begin{aligned}
\mathcal{I}_{m_1}^t &= \big\{ i \in \{1,\dots,\mathcal{B}\} \ \big| \ D_{m_2,i}^t < D_{m_1,i}^t \big\}, \\
\mathcal{I}_{m_2}^t &= \big\{ i \in \{1,\dots,\mathcal{B}\} \ \big| \ D_{m_1,i}^t  \le D_{m_2,i}^t \big\}.
\end{aligned}
\end{equation}
Here, $\mathcal{I}_{m_1}^t$ is the index set of samples within the $t$-th batch that are determined to exhibit distribution shift in modality $m_1$, and the same applies to the notation $\mathcal{I}_{m_2}^t$.

During the pre-training process of the backbone model \cite{gong2023contrastive}, the alignment of different uni-modal features is conducted via contrastive training \cite{oord2018representation}. Therefore, we align the less reliable view toward the reliable one using a \emph{one-sided} InfoNCE loss: only the \emph{less reliable} side receives gradients, while the \emph{reliable} side is stop-grad to avoid mutual dragging.
Let $\hat{\boldsymbol{z}}_{m_1,i}^t=\boldsymbol{z}_{m_1,i}^t/\|\boldsymbol{z}_{m_1,i}^t\|_2$  and $\hat{\boldsymbol{z}}_{m_2,i}^t=\boldsymbol{z}_{m_2,i}^t/\|\boldsymbol{z}_{m_2,i}^t\|_2$ 
be $\ell_2$-normalized features and $\tau=0.05$ be the temperature.
We denote stop-gradient by $\operatorname{sg}(\cdot)$ and inner product similarity by $\langle \cdot,\cdot\rangle$. For each sample $i$ in batch $t$, the loss is defined as
\begin{equation}
\label{eq:onesided_infonce_instance}
\ell_i =
\begin{cases}
-\log \dfrac{\exp\big(\langle \hat{\boldsymbol{z}}_{m_1,i}^t,\operatorname{sg}(\hat{\boldsymbol{z}}_{m_2,i}^t)\rangle / \tau\big)}
{\sum_{j=1}^\mathcal{B} \exp\big(\langle \hat{\boldsymbol{z}}_{m_1,i}^t,\operatorname{sg}(\hat{\boldsymbol{z}}_{m_2,j}^t)\rangle / \tau\big)}, 
& i \in \mathcal{I}_{m_1}^t, \\[12pt]
-\log \dfrac{\exp\big(\langle \hat{\boldsymbol{z}}_{m_2,i}^t,\operatorname{sg}(\hat{\boldsymbol{z}}_{m_1,i}^t)\rangle / \tau\big)}
{\sum_{j=1}^\mathcal{B} \exp\big(\langle \hat{\boldsymbol{z}}_{m_2,i}^t,\operatorname{sg}(\hat{\boldsymbol{z}}_{m_1,j}^t)\rangle / \tau\big)}, 
& i \in \mathcal{I}_{m_2}^t.
\end{cases}
\end{equation}

The overall alignment loss for batch $t$ is then obtained by averaging over all samples:
\begin{equation}
\mathcal{L}_{\text{c}} = \frac{1}{\mathcal{B}}\sum_{i=1}^\mathcal{B} \ell_i.
\end{equation}


\begin{table*}[ht]
  \caption{
  Prediction accuracies (in \%) on Kinetics50-C benchmark under corrupted video modality. 
  The accuracy values for each corruption type are presented, along with the average accuracy. 
  Best results are highlighted in bold and second-best results are underlined. 
  }
  \label{tab:ks50}
  \centering
\scriptsize
\begin{adjustbox}{width=\textwidth}
\begin{tabular}{lccccccccccccccccc} 
\toprule
\multirow{2}{*}{Models} & \multicolumn{3}{c}{Noise} & \multicolumn{4}{c}{Blur} & \multicolumn{4}{c}{Weather} & \multicolumn{4}{c}{Digital} & \multirow{2}{*}{Avg.} \\
\cmidrule(lr){2-4} \cmidrule(lr){5-8} \cmidrule(lr){9-12} \cmidrule(lr){13-16}
& Gauss. & Shot & Impul. & Defoc. & Glass & Mot. & Zoom & Snow & Frost & Fog & Brit. & Contr. & Elas. & Pix. & JPEG \\
\midrule
Source & 46.8 & 48 & 46.9 & 67.5 & 62.2 & 70.8 & 66.7 & 61.6 & 60.3 & 46.7 & 75.2 & 52.1 & 65.7 & 66.5 & 61.9 & 59.9 \\ 
Tent & 46.3 & 47 & 46.3 & 67.2 & 62.5 & 71 & 67.6 & 63.1 & 61.1 & 34.9 & 75.4 & 51.6 & 66.8 & 67.2 & 62.7 & 59.4 \\ 
EATA & 46.8 & 47.6 & 47.1 & 67.2 & 62.7 & 70.6 & 67.2 & 62.3 & 60.9 & 46.7 & 75.2 & 52.4 & 65.9 & 66.8 & 62.5 & 60.1 \\ 
SAR & 46.7 & 47.4 & 46.8 & 67 & 61.9 & 70.4 & 66.4 & 61.8 & 60.6 & 46 & 75.2 & 52.1 & 65.7 & 66.4 & 62 & 59.8 \\ 
MMT & 46.2 & 46.6 & 46.1 & 58.8 & 55.7 & 62.4 & 61.7 & 52.6 & 54.4 & 48.5 & 69.3 & 49.3 & 57.6 & 56.4 & 54.5 & 54.5 \\ 
READ & 49.4 & 49.7 & 49 & 68 & 65.1 & 71.2 & 69 & 64.5 & 64.4 & 57.4 & 75.5 & 53.6 & 68.3 & 68 & 65.1 & 62.5 \\ 
SuMi & 50.1 & \underline{50.7} & \underline{50.4} & 68.2 & 65.6 & \underline{72.2} & \underline{69.7} & \underline{65.7} & \underline{67.0} & 56.5 & \textbf{77.1} & 55.2 & 69.3 & \underline{71.2} & \underline{68.9} & 63.9 \\ 
TSA & \textbf{52.6} & \textbf{52.3} & \textbf{52} & \underline{68.7} & \underline{68} & 70.7 & 68.8 & 65.2 & 66.6 & \underline{64.3} & 74.6 & \underline{57.4} & \underline{70.5} & 69 & 66.2 & \underline{64.5} \\ 
\rowcolor{blue!20} \textbf{AdaPGC} & \underline{50.6} & 50.6 & 50.2 & \textbf{70.7} & \textbf{70.8} & \textbf{72.5} & \textbf{72.3} & \textbf{67.5} & \textbf{68.6} & \textbf{67.8} & \underline{76.3} & \textbf{58.7} & \textbf{72.9} & \textbf{72.0} & \textbf{69.6} & \textbf{66.1} \\
\bottomrule
\end{tabular}
 \end{adjustbox}
\end{table*}


\begin{table*}[ht]
  \caption{Prediction accuracies (in \%) on Kinetics50-C (left) and VGGSound-C (right) benchmarks under corrupted audio modality. 
  The accuracy values are shown for each corruption type, with the average accuracy for each model across the benchmarks. 
  Best results are highlighted in bold and second-best results are underlined.
  }
  \label{tab:audio-C}
  \centering
\scriptsize
\begin{adjustbox}{width=\textwidth}
\begin{tabular}{lcccccccccccccccc} 

\toprule
\multirow{2}{*}{Models} & \multicolumn{3}{c}{Noise} & \multicolumn{3}{c}{Weather} & \multirow{2}{*}{Avg.}  & \multicolumn{3}{c}{Noise} & \multicolumn{3}{c}{Weather} & \multirow{2}{*}{Avg.}  \\
\cmidrule(lr){2-4} \cmidrule(lr){5-7} \cmidrule(lr){9-11} \cmidrule(lr){12-14}
& Gauss. & Traff. & Crowd. & Rain & Thund. & Wind & & Gauss. & Traff. & Crowd. & Rain & Thund. & Wind \\
\midrule
Source & 73.7 & 65.5 & 67.9 & 70.3 & 67.9 & 70.3 & 69.3 & 37 & 25.5 & 16.8 & 21.6 & 27.3 & 25.5 & 25.6 \\ 
Tent & 73.9 & 67.4 & 69.2 & 70.4 & 66.5 & 70.5 & 69.6 & 10.6 & 2.6 & 1.8 & 2.8 & 5.3 & 4.1 & 4.5 \\ 
EATA & 73.7 & 66.1 & 68.5 & 70.3 & 67.9 & 70.1 & 69.4 & 39.2 & 26.1 & 22.9 & 26 & 31.7 & 30.4 & 29.4 \\ 
SAR & 73.7 & 65.4 & 68.2 & 69.9 & 67.2 & 70.2 & 69.1 & 37.4 & 9.5 & 11 & 12.1 & 26.8 & 23.7 & 20.1 \\ 
MMT & 70.8 & 69.2 & 68.5 & 69 & 69.8 & 68.5 & 69.4 & 14.1 & 5.2 & 6.4 & 9.8 & 8.6 & 4.5 & 7.6 \\ 

READ & 74.1 & 69.0 & 69.7 & 71.1 & 71.8 & 70.7 & 71.1 & 40.4 & 28.9 & 26.6 & 30.9 & 36.7 & 30.6 & 32.4 \\

SuMi & \underline{75.1} & 68.9 & \underline{70.6} & \underline{71.6} & \underline{72.8} & \underline{72.1} & \underline{71.9} & \textbf{41.9} & 26.3 & 27.9 & 31.6 & 37.1 & \underline{34.1} & 33.2 \\

TSA & 74.5 & \underline{69.6} & 70.5 & 71.4 & 72.0 & 71.0 & 71.5 & \underline{41.5} & \underline{31.8} & \underline{30.9} & \textbf{32.6} & \underline{38.9} & 32.6 & \underline{34.7} \\

\rowcolor{blue!20} \textbf{AdaPGC} & \textbf{75.5} & \textbf{71.8} & \textbf{72.1} & \textbf{72.0} & \textbf{75.0} & \textbf{72.7} & \textbf{73.2} & 40.4 & \textbf{35.5} & \textbf{36.1} & \underline{31.9} & \textbf{40.2} & \textbf{36.4} & \textbf{36.8} \\
\bottomrule

\end{tabular}
\end{adjustbox}
\end{table*}

\subsection{Overall Optimization}

\paragraph{Regularization for Stable Adaptation.}

To stabilize test-time optimization and avoid prediction degeneration, we follow Yang et al.~\cite{DBLP:conf/iclr/Yang0Z0024} and introduce two lightweight entropy-based regularizers.  
Given $p_i=\text{Softmax}(s(z_{F,i}))$, let $u_i=\max_c p_{i,c}$ and
$q=\text{Softmax}\!\Big(\sum_{i=1}^{\mathcal{B}} p_i\Big).$
The confidence and balance regularizers are
\begin{equation}
    \mathcal{L}_{\mathrm{ra}} = -\tfrac{1}{\mathcal{B}}\sum_{i} u_i \log u_i,
\quad
\mathcal{L}_{\mathrm{bal}} = -\sum_{c} q_c \log q_c.
\end{equation}
The former prevents overly peaked predictions, while the latter encourages batch-level category balance.

\paragraph{Overall objective and inference.}
The overall loss is
\begin{equation}
\mathcal{L}
=  \mathcal{L}_{\text{ra}}
+ \mathcal{L}_{\text{bal}}
+ w_c\,\mathcal{L}_{\text{c}}
+ w_g\,\mathcal{L}_{\text{g}},
\end{equation}
where only $\mathcal{L}_{\text{g}}$ and $\mathcal{L}_{\text{c}}$ are weighted by hyperparameters $(w_g,w_c)$; the two regularizers $ \mathcal{L}_{\mathrm{ra}},  \mathcal{L}_{\mathrm{bal}}$ are unweighted.

%% file: sec/4_exp.tex
\section{Experiments}

\begin{table*}[ht]
  \caption{Prediction accuracies (in \%) on VGGSound-C benchmark under corrupted video modality. 
  The accuracy values for each corruption type are presented, along with the average accuracy. 
  Best results are highlighted in bold and second-best results are underlined. 
  }
  \label{tab:vgg}
  \centering
\scriptsize
\begin{adjustbox}{width=\textwidth}
\begin{tabular}{lccccccccccccccccc}
\toprule
\multirow{2}{*}{Models} & \multicolumn{3}{c}{Noise} & \multicolumn{4}{c}{Blur} & \multicolumn{4}{c}{Weather} & \multicolumn{4}{c}{Digital} & \multirow{2}{*}{Avg.} \\
\cmidrule(lr){2-4} \cmidrule(lr){5-8} \cmidrule(lr){9-12} \cmidrule(lr){13-16}
& Gauss. & Shot & Impul. & Defoc. & Glass & Mot. & Zoom & Snow & Frost & Fog & Brit. & Contr. & Elas. & Pix. & JPEG \\
\midrule
Source & 52.8 & 52.7 & 52.7 & 57.2 & 57.2 & 58.7 & 57.6 & 56.4 & 56.6 & 55.6 & 58.9 & 53.7 & 56.9 & 55.8 & 56.9 & 56 \\ 
Tent & 52.7 & 52.7 & 52.7 & 56.7 & 56.5 & 57.9 & 57.2 & 55.9 & 56.3 & 56.3 & 58.4 & 54 & 57.4 & 56.2 & 56.7 & 55.8 \\ 
EATA & 53 & 52.8 & 53 & 57.2 & 57.1 & 58.6 & 57.8 & 56.3 & 56.8 & 56.4 & 59 & 54.1 & 57.4 & 56.1 & 57 & 56.2 \\ 
SAR & 52.9 & 52.8 & 52.9 & 57.2 & 57.1 & 58.6 & 57.6 & 56.3 & 56.7 & 55.9 & 58.9 & 54 & 57 & 56 & 57 & 56.1 \\ 
MMT & 7.1 & 7.3 & 7.3 & 44.8 & 41.5 & 48 & 45.5 & 27.4 & 23.5 & 30.5 & 46.3 & 24 & 43 & 40.7 & 45.7 & 32 \\ 
READ & 53.6 & 53.6 & 53.5 & \underline{57.9} & 57.7 & \textbf{59.4} & \textbf{58.8} & \underline{57.2} & \underline{57.8} & 55 & \underline{59.9} & \textbf{55.2} & \underline{58.6} & 57.1 & 57.9 & 56.9 \\

SuMi & \textbf{54} & \underline{54.3} & 53.8 & \textbf{58.2} & \textbf{58.4} & \textbf{59.4} & \underline{58.7} & \textbf{57.5} & \textbf{58.2} & \underline{57.6} & 59.4 & 54.8 & \textbf{59.0} & \underline{57.5} & \textbf{58.2} & \textbf{57.3} \\

TSA & \underline{53.9} & 53.9 & \underline{54} & 57.6 & \underline{58} & \underline{59} & 58.6 & 56.9 & 57 & 56.6 & 59.8 & 54.7 & \underline{58.6} & 56.7 & \underline{57.9} & 56.9 \\

\rowcolor{blue!20}
\textbf{AdaPGC} & \underline{53.9} & \textbf{54.4} & \textbf{54.2} & 57.1 & 57.0 & 58.6 & 58.3 & 57.1 & 57.2 & \textbf{58.0} & \textbf{60.0} & \underline{55.1} & \textbf{59.0} & \textbf{57.6} & \textbf{58.2} & \underline{57.0} \\
\bottomrule
\end{tabular}
\end{adjustbox}
\end{table*}

\begin{table*}[t]
\centering
\caption{
Ablation study on the contributions of different components in AdaPGC on the Kinetics50-C benchmark. 
For the video branch, we only display a representative subset of corruption types (Defocus, Glass, Snow, Frost, Elastic, Pixelate) for clarity. 
}
\label{tab:contribution-ab}
\scriptsize
\begin{adjustbox}{width=\textwidth}
\begin{tabular}{ccc|ccccccc|ccccccc}
\toprule
& & & \multicolumn{7}{c|}{\textbf{Video-C}} & \multicolumn{7}{c}{\textbf{Audio-C}} \\
\textbf{FL} & \textbf{PA} & \textbf{AR} &
Defoc. & Glass & Snow & Frost & Elas. & Pix. & Avg. & Gauss. & Traff. & Crowd. & Rain & Thund. & Wind & Avg.\\ 
\midrule
 &  &  & 68.39 & 66.41 & 65 & 66.21 & 69.64 & 68.1 & 63.81 & 74.48 & 70.73 & 70.56 & 71.45 & 72.9 & 71.13 & 71.88 \\ 
\checkmark &  &  & 68.19 & 67.1 & 64.84 & 67.02 & 69.4 & 68.63 & 63.57 & 74.52 & 71.01 & 71.17 & 71.01 & 74.03 & 71.85 & 72.27 \\ 
 & \checkmark &  & 68.95 & 67.62 & 65.24 & 66.73 & 70.04 & 69.64 & 63.94 & 74.23 & 70.97 & 70.44 & 71.81 & 73.35 & 71.65 & 72.08 \\ 
 &  & \checkmark & 69.19 & 68.06 & 66.49 & 67.3 & 70.52 & 70.73 & 65.09 & 74.84 & 70.65 & 71.05 & 72.18 & 73.35 & 71.45 & 72.25 \\ 
\checkmark & \checkmark &  & 69.56 & 69.48 & 66.09 & 67.98 & 71.98 & 70.65 & 64.9 & 75.08 & 71.29 & 72.14 & 72.26 & 74.52 & 72.06 & 72.90 \\ 
\checkmark &  & \checkmark & 69.31 & 69.68 & 67.1 & 67.86 & 71.94 & 71.13 & 65.16 & 75.28 & 71.21 & 71.98 & 71.81 & 75.16 & 72.06 & 72.92 \\ 
 & \checkmark & \checkmark & 69.92 & 69.31 & 66.29 & 67.66 & 71.05 & 71.29 & 64.93 & 74.72 & 71.41 & 70.93 & 71.9 & 73.83 & 72.06 & 72.48 \\ 
\checkmark & \checkmark & \checkmark & \textbf{70.65} & \textbf{70.77} & \textbf{67.54} & \textbf{68.63} & \textbf{72.94} & \textbf{71.98} & \textbf{66.08} & \textbf{75.48} & \textbf{71.81} & \textbf{72.1} & \textbf{71.98} & \textbf{75.04} & \textbf{72.7} & \textbf{73.19} \\ 
\bottomrule
\end{tabular}
\end{adjustbox}
\end{table*}
\subsection{Experimental Settings}

\paragraph{Datasets.}

We evaluated on two corrupted datasets, Kinetics50-C and VGGSound-C, derived from Kinetics50 \cite{kay2017kinetics} and VGGSound \cite{chen2020vggsound}. Kinetics50 contains YouTube videos of human actions across 50 categories, while VGGSound includes 309 audio-visual classes, covering a wide range of real-world audio-visual events. Following previous work \cite{hendrycksbenchmarking,DBLP:conf/iclr/Yang0Z0024}, we introduced 15 visual distortions (e.g., Gaussian noise, motion blur, brightness variations) and six audio disturbances (e.g., traffic noise, rain noise) to simulate real-world distribution shifts. These distortions resulted in the corrupted Kinetics50-C and VGGSound-C datasets, which were then used for our evaluation, providing a challenging benchmark for multi-modal test-time adaptation.

\paragraph{Implementation Details.}
For the source model, we follow READ \cite{DBLP:conf/iclr/Yang0Z0024} and adopt the pre-trained CAV-MAE \cite{gong2023contrastive}. 
All experiments use the Adam optimizer with a learning rate of $1\times10^{-4}$ and a batch size of 16 for both the Kinetics50-C and VGGSound-C datasets. 
We configure the hyperparameters using loss weights $w_c=0.01$ and $w_g=1$, together with a fusion weight $\lambda=1$.
We use two NVIDIA RTX 4090 GPUs to optimize processing efficiency. One GPU is dedicated to storing the source model and updating its parameters, while the other handles the GDA models, including storing and updating their parameters as well as making predictions. 
During adaptation, we enforce a clear separation of update routes. 
The losses $\mathcal{L}_{\text{g}}$ together with $\mathcal{L}_{\text{ra}}$ and 
$\mathcal{L}_{\text{bal}}$ update \emph{only} the fusion-layer attention parameters 
$W_{\Theta_h}$ and $B_{\Theta_h}$ for $h \in \{Q, K, V\}$, without modifying any 
modality-specific encoders. In contrast, $\mathcal{L}_{\text{c}}$ updates 
\emph{only} the LayerNorm modules within the encoder of the modality flagged as 
corrupted by the reliability test.

\subsection{Comparison with the State-of-the-Art Methods}
We compare the proposed method AdaPGC with several existing approaches, including uni-modal TTA methods such as TENT \cite{wang2021tent}, EATA \cite{pmlr-v162-niu22a}, and SAR \cite{niu2023towards}, as well as multi-modal TTA methods like MM-TTA \cite{shin2022mm}, READ \cite{DBLP:conf/iclr/Yang0Z0024}, SuMi \cite{guo2025smoothing}, and TSA \cite{chen2025testtime}.

The performance of AdaPGC is evaluated against these baselines, as shown in Table \ref{tab:ks50}, Table \ref{tab:audio-C} and Table \ref{tab:vgg}, where the Source model represents the baseline without any test-time adaptation. From the results, we observe the following:
\begin{itemize}
\item 
AdaPGC demonstrates exceptional robustness across various corruption settings, outperforming all existing uni-modal TTA methods and most existing multi-modal TTA methods. 
This highlights its enhanced adaptability in multi-modal scenarios, as AdaPGC can more effectively exploit information from the unimpaired modality to compensate for the degraded one.

\item READ, SuMi and TSA have certain limitations. 
Although READ constrains output predictions to maintain partial stability under modality corruption, 
SuMi enhances robustness mainly through smoothing and cross-modal information sharing, 
and TSA selects unimpaired modalities for model correction, 
none of these methods can continuously track feature distributions across modalities.This limits their ability to accurately characterize the evolving distribution of corrupted modalities, which may explain their inferior performance under challenging corruption settings.
In contrast, AdaPGC integrates an incremental GDA mechanism, continuously updating category statistics and adaptively correcting decision boundaries at each test stage by modeling the underlying distribution of corrupted modalities.

\item 
When subjected to some complex perturbations, such as fog, pixelate and wind, AdaPGC shows clear performance gains. Its average accuracy exceeds that of most other methods, suggesting the effectiveness of AdaPGC’s incremental updating strategy and its ability to recover from perturbations in dynamic environments.

\end{itemize}

\subsection{Ablation Study}
\paragraph{Contributions of different components.}
We conduct an ablation study (Table~\ref{tab:contribution-ab}) to examine the individual contributions of AdaPGC’s three principal components: 
the fused logits mechanism (FL; \cref{eq:final-pre}), 
the prediction alignment loss (PA; \cref{eq:l-g}), 
and the asymmetry rectification module (AR).
Starting from the baseline without these modules, each component improves the average accuracy on both Video-C and Audio-C, showing that all three are effective.
Among them, FL+PA corresponds to the canonical GDA setting discussed in \textbf{Section} \ref{sec:intro}, as FL incorporates the GDA posterior into prediction fusion and PA aligns the backbone predictions with the GDA-based predictions.
Adding AR further improves accuracy, demonstrating that rectifying modality asymmetry is crucial in multi-modal TTA beyond canonical GDA.


\paragraph{Hyperparameters Sensitivity Analysis.}

\begin{figure}
  \centering
  \includegraphics[width=1.01\linewidth]{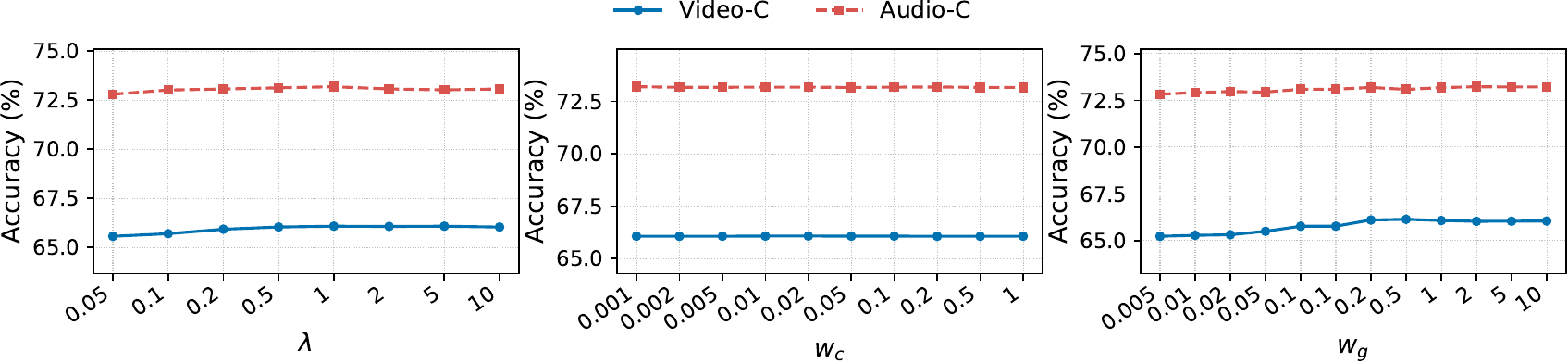}
\caption{
Hyperparameter sensitivity visualization for AdaPGC on the Kinetics50-C benchmark. 
Each plot shows the average model accuracy under different values of the corresponding hyperparameter for the Video-C and Audio-C tasks.
}
  \label{fig:hyper}
\end{figure}

We further analyze the sensitivity of AdaPGC to its major hyperparameters, including the fusion coefficient $\lambda$, the contrastive consistency loss weight $w_c$, and the prediction alignment loss weight $w_g$. 
As shown in Figure~\ref{fig:hyper}, AdaPGC is insensitive to the choices of $\lambda$, $w_c$, and $w_g$. 
The accuracy remains stable across all tested values, with slight improvements near $\lambda=1$ and $w_g=1$. 
Overall, the relatively flat trends indicate that our method is robust and does not rely on careful hyperparameter tuning.

%% file: sec/5_conclusion.tex
\section{Conclusion}

This work introduces AdaPGC, a novel multi-modal test-time adaptation framework designed to achieve reliable predictions under uni-modal distribution shifts. 
Instead of depending solely on network outputs, AdaPGC incorporates an explicit probabilistic view of category-wise feature distributions and updates these statistics online without requiring labeled data. 
Moreover, by identifying modality-specific degradation and applying a contrastive rectification strategy, the framework effectively alleviates the imbalance caused by asymmetric modality shifts. 
Through experiments on two representative multi-modal benchmarks, AdaPGC consistently delivers stronger robustness and more stable decision boundaries compared with existing TTA approaches, demonstrating its practicality for real-world multi-modal systems.